\documentclass[letterpaper, 10 pt, conference]{ieeeconf}  

\IEEEoverridecommandlockouts                              

\usepackage{comment}

\usepackage{glossaries}
\makeglossaries

\newacronym{lfd}{LfD}{Learning from Demonstration}
\newacronym{vae}{VAE}{Variational Auto-Encoder}
\newacronym{kl}{KL}{Kullback–Leibler}
\newacronym{ssl}{SSL}{Self-Supervised Loss}
\newacronym{rmse}{RMSE}{Root Mean Squared Error}
\newacronym{pca}{PCA}{Principal Component Analysis}
\newacronym{dof}{DoF}{Degrees of Freedom}
\newacronym{tsne}{t-SNE}{t-distributed Stochastic Neighbor Embedding}
\newacronym{mae}{MAE}{masked autoencoder}
\newacronym{byol}{BYOL}{Bootstrap Your Own Latent}
\newacronym{relu}{ReLU}{Rectified Linear Unit}

\usepackage{amsmath}
\usepackage{amsfonts}

\usepackage{siunitx}

\usepackage{graphicx}
\usepackage{multirow}
\usepackage{hhline}
\usepackage{adjustbox}
\usepackage{tabularray}

\usepackage{subcaption}

\usepackage[backend=biber,style=ieee,mincitenames=1,maxnames=20,maxcitenames=1,natbib=true]{biblatex}
\usepackage{doi}

\usepackage{xcolor}

\usepackage[capitalise]{cleveref}
\usepackage{nameref}
\hypersetup{%
 setpagesize=false,%
 bookmarks=true,%
 bookmarksdepth=tocdepth,%
 bookmarksnumbered=true,%
 colorlinks=true,%
 linkcolor=blue,
 citecolor=blue,
 urlcolor=blue,
 pdftitle={},%
 pdfsubject={},%
 pdfauthor={},%
 pdfkeywords={}}

\newcommand{\paraDraft}[1]{\ifdefined\draft\subsubsection*{\color{blue}\textbf{#1}}\fi}

\title{\LARGE \bf
Latent Object Characteristics Recognition with\\ Visual to Haptic-Audio Cross-modal Transfer Learning 
}

\author{Namiko Saito$^{1}$, Jo\~{a}o Moura$^{1}$, Hiroki Uchida$^{2}$, and Sethu Vijayakumar$^{1}$
\thanks{$^{1}$Authors are with the School of Informatics, The University of Edinburgh, Edinburgh, U.K., and with The Alan Turing Institute, London, U.K.}
\thanks{$^{2}$Author is with the Department of Intermedia Art and Science, Waseda University, Tokyo, Japan}%
}

\begin{document}

\maketitle
\thispagestyle{empty}
\pagestyle{empty}

\begin{abstract}
Recognising the characteristics of objects while a robot handles them is crucial for adjusting motions that ensure stable and efficient interactions with containers. 
Ahead of realising stable and efficient robot motions for handling/transferring the containers, this work aims to recognise the latent unobservable object characteristics.
While vision is commonly used for object recognition by robots, it is ineffective for detecting hidden objects.
However, recognising objects indirectly using other sensors is a challenging task. 
To address this challenge, we propose a cross-modal transfer learning approach from vision to haptic-audio. 
We initially train the model with vision, directly observing the target object. 
Subsequently, we transfer the latent space learned from vision to a second module, trained only with haptic-audio and motor data. 
This transfer learning framework facilitates the representation of object characteristics using indirect sensor data, thereby improving recognition accuracy.
For evaluating the recognition accuracy of our proposed learning framework we selected shape, position, and orientation as the object characteristics.
Finally, we demonstrate online recognition of both trained and untrained objects using the humanoid robot Nextage Open.
\end{abstract}

\section{INTRODUCTION}
\label{s:introduction}
\paraDraft{Robot motion depending on object characteristics}
Interacting with containers is a fundamental task in robotics, applicable to warehouse, delivery, and home use.
These scenarios require robots to handle containers efficiently while keeping their content safe and stable, i.e. without having the objects topple, mix, and get damaged through impact.
Therefore, robots need to adjust their motion (speed, acceleration, and trajectory) to the specific characteristics of the contained objects.
However, often these objects are hidden by the container's cover; therefore, ahead of realizing stable and quick robot motions for handling/transferring the containers, this work aims to recognise the latent unobservable object characteristics.

\paraDraft{predicting latent characteristics is difficult, so try cross-modal}
Recognising hidden object characteristics inside a closed container presents a significant challenge.
One solution involves leveraging indirect sensing modalities such as haptic and/or audio cues to infer these hidden characteristics as~\cite{Chen2021, Niwa2007}.
However, we will show that simply using haptic and audio as indirect sensing fails to accomplish good recognition.
To enable using haptic-audio as indirect sensing to estimate the hidden characteristics, in this work we propose a cross-modal transfer learning framework, where we first learn a prediction model based on vision reconstruction, which can directly observe the target characteristics, and transfer its latent space to a second prediction model learnt only with haptic-audio sensing information.
This approach aims to bridge the gap between direct and indirect sensing modalities, facilitating more accurate recognition of hidden latent object characteristics using haptic and audio cues.

\paraDraft{Object behavior}
There are various latent characteristics that influence object behaviour as addressed by~\citet{Gao2023}, who focus on factors such as mass, fragility, deformability, and certain properties that humans can estimate from objects' appearance.
~\citet{Funabashi2022} examine attributes like heaviness, softness, and slipperiness, which are related to the sense of touch.
Additionally~\citet{Gonçalves2014} categorize objects based on characteristics such as area, squareness, circleness, and other geometric properties.
As an exemplar, in this work, we focus on recognising the shape, orientation, and position of the objects within the container to assess our transfer learning framework, as they provide measurable ground truth that allows for numerical evaluation of accuracy.

\begin{figure}[t]
    \centerline{\includegraphics[width=8cm]{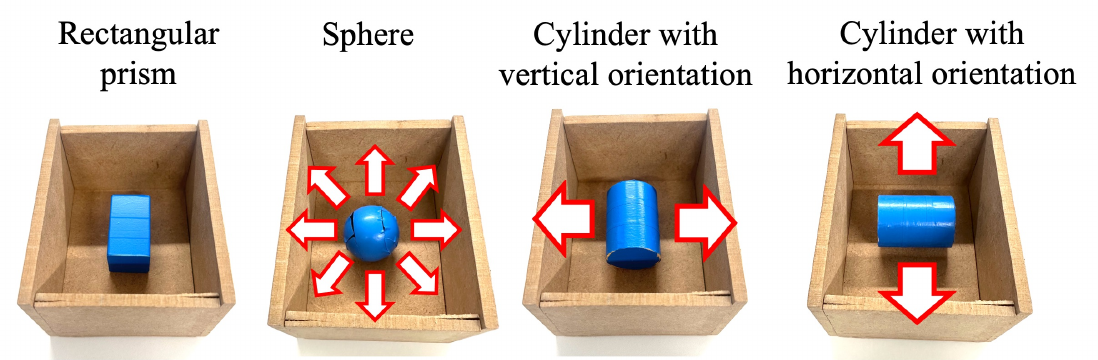}}
    \caption{
        A rectangular prism typically offers greater stability compared to a cylinder or a sphere. 
        The ease with which a cylinder rolls is contingent upon its orientation, while a sphere possesses the capability to roll in any direction.
    }
    \label{f:motivation}
\end{figure}

\paraDraft{example of shape, orientation and position}
Shape, orientation, and position are fundamental attributes that influence object behavior and are often concealed by the container's cover.
For example, as shown in~\cref{f:motivation}, shape and orientation affect the direction of motion. 
While rectangular-shaped objects are stable, spheres can easily move in every direction.
For the cylinder, in addition to the shape, one has to consider the orientation when predicting the direction of its rolling motion. 
Moreover, the position of the object within the container also influences its motion. 
Objects positioned at the centre of the container have more freedom of movement, while those placed near the corners are constrained by the container's walls, limiting their motion options.

\begin{figure*}[t]
    \centerline{\includegraphics[width=18cm]{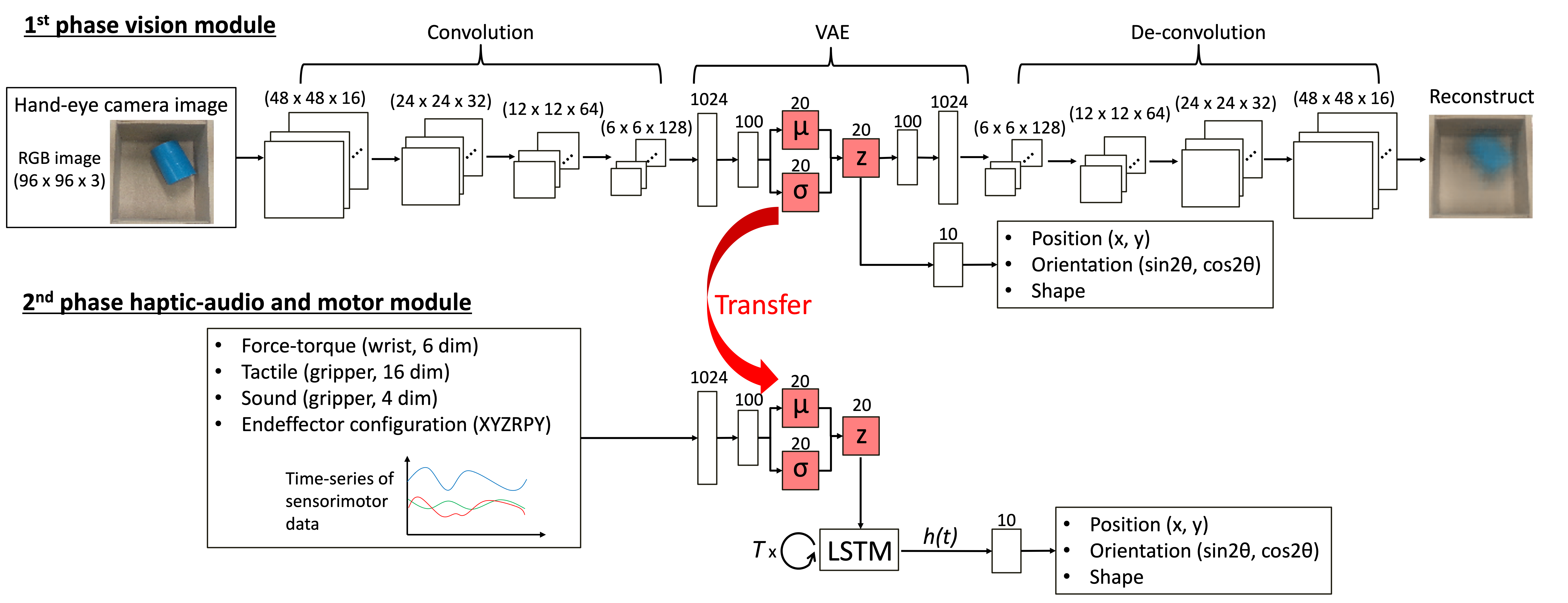}}
    \caption{
       Overall learning model.
       We train the first phase vision module, and then transfer the latent space to the second phase haptic-audio and motor module to use it for initiating the training process of the latter module. 
       Our objective is to discern object position, orientation, and shape utilising indirect information derived from haptic-audio cues.
    }
    \label{f:model}
\end{figure*}
\section{Related Work}
\subsection{Object characteristics recognition with vision, haptics and audio}
\paraDraft{Using vision}
There are several methods for object detection~\cite{Redmon2016, Liu2016, Padilla2020}, and recognition~\cite{Loncomilla2016, Eitel2015, Maturana2015} with vision.
In contrast,~\citet{Gao2023} addressed the understanding of physical concepts such as mass, fragility, and deformability for robot manipulation employing LLM and relied on human annotations based on vision to achieve this understanding.

\paraDraft{Using tactile}
Using GelSight~\cite{Yuan2017} vision-tactile sensors,~\citet{Anzai2020} achieved perception of object pose within the hand,~\citet{Lin2019} achieved object identification from data samples through object picking, while~\citet{Jiang2022} realized the perception of material, shape, and pose of objects. 
~\citet{Falco2017} and~\citet{Li2019} employed cross-modal learning, integrating tactile and vision, for recognising object shapes.

\paraDraft{Using Audio}
~\citet{Sterling2018} utilised impact sound data generated during hitting an object to estimate the object's geometry and material properties, and also reconstructed image data. 
~\citet{Chen2021} introduced a learning model to predict object shape by analysing acoustic vibrations captured when dropping an object into a container. 
Additionally,~\citet{Niwa2007} employed acoustic diffraction of audible sound to estimate the distance of an object, even when occluded by another object.

\paraDraft{Problems}
The aforementioned research successfully recognised static object characteristics. 
However, challenges remain in recognising characteristics dynamically as robots perform tasks. 
Our objective is to address this gap by developing a system capable of identifying object characteristics on the fly, while the robot is executing a task. 
This online recognition is crucial for generating targeted motions that depend on the object characteristics in future stages of the task.

\subsection{Dynamic object characteristics recognition during motions}
\paraDraft{Tracking with vision or tactile}
\citet{Marturi2019} demonstrates a robot arm tracking and grasping moving objects using a vision system.
Additionally,~\citet{Chen2023} achieved 6D pose dynamic estimation through optical flow analysis while manipulating objects.
\citet{Funabashi2022} used uSkin~\cite{Tomo2018} skin-type sensors which can detect force in 3-axis, to recognise heaviness, softness, and slipperiness of an object while a multi-fingered hand grasps it.

\paraDraft{Multimodal dynamic representation}
\citet{Saito2021} and~\citet{Gemici2014} represent the characteristics of objects using the neural networks' latent space values. 
These networks utilise multimodal data which includes vision, tactile, and force sensing collected while the robot arm is in motion.

\paraDraft{Problems}
Previous work relies on sensing modalities such as vision, touch, or vibration that can directly observe the target object characteristics.
In contrast, this work focuses on the scenario of hidden target characteristics, for instance the case of recognizing visual information, such as position and shape of an object inside of a container, where the robot can only indirectly sense these characteristics.
This indirect sensing presents additional challenges in prediction and recognition.

\section{Approach}
\paraDraft{Objective and task}
The goal of this work is to achieve recognition of object characteristics while in motion, focusing particularly on the case of indirect sensing of those characteristics.
To address this challenge, we propose a cross-modal transfer learning method.
We choose a box transfer task as an example to evaluate our method. 
In this task, a humanoid robot, Nextage Open from Kawada Robotics, grasps a box containing a small object and moves the box to different poses, randomly generated. 
As the robot moves, our learning model predicts the characteristics of the small object, such as its shape, position, and orientation, on the fly.

\paraDraft{Cross modal approach}
Previous research on cross-modal learning~\cite{Lin2019, Falco2017, Li2019}, which introduced learning models trained with both vision and tactile data to predict object class or shape, addressed occlusion and variable lighting conditions. 
Inspired by these studies, we propose a two-phase cross-modal transfer learning approach. 
Initially, we train a vision module by opening the lid of the container to enable direct visual observation of the object. 
Subsequently, we transfer the learned latent space to the second phase, where we train the model using haptic and audio data with the container closed, allowing the robot to sense the object indirectly.
We hypothesise that warm starting the training of the second phase, which uses haptic and audio data, with the latent space of the visual module will improve the recognition of the object characteristics.

\paraDraft{contribution}
In summary, our contributions are as follows:
\begin{itemize} 
    \item Proposing a two-phase learning framework for recognising latent object characteristics.
    \item Demonstrating that warm starting the initial haptic-audio latent state with the learnt visual latent state significantly improves the recognition of visual characteristics when using only indirect haptic-audio and motor sensing.
    \item Validating the proposed learning framework on online prediction of object shape, position, and orientation on a physical humanoid robot setup.
\end{itemize}

\section{Method}
\paraDraft{Overall}
\cref{f:model} depicts the overall architecture of the learning modules. 
We train the vision module in the first phase and transfer its latent space to the second phase haptic-audio and motor module. 
This transfer serves as the initial state value, improving the training convergence of the second phase module.
Subsequently, as a baseline, we train the second phase to recognise object position, orientation, and shape, without using the warm start from the visual module latent space.

\subsection{1st phase vision module}
This module aims to predict object characteristics from RGB images in situations where objects are directly observable within the container. 
To minimize the cost of collecting training data, we captured snapshot images of objects randomly placed inside the box, without moving the robot.

The structure of the module consists of convolution layers, variational autoencoder (VAE)~\cite{Kingma2013} and de-convolution layers, with the number of neurons detailed in~\cref{f:model}.
In the convolutional layers, we utilise a kernel size of $4\times4$, a stride of $2$, and zero padding of $1$. 
These layers convolute the image data and reconstruct it ($y_{{\rm {image}}}$). 
Additionally, we incorporate a fully connected layer from the latent space of the VAE to output continuous values representing position and orientation ($y_{{\rm {pos}}}$, $y_{{\rm {ori}}}$) and shape labels ($y_{{\rm {shape}}}$). 

The architecture uses a self-supervised approach for image reconstruction and supervised learning with ground truth data for characteristics prediction.
The loss function $L$ is 
\begin{equation}
L = \alpha{\rm {D_{{\rm KL}}}}({\rm P}||{\rm Q}) + \beta{\rm {E_{{\rm reconst}}}} + {\rm {E_{{\rm predict}}}},
\label{1st_loss}
\end{equation}
where
\begin{equation}
{\rm {D_{{\rm KL}}}}({\rm P}||{\rm Q}) 
 = - \frac{1}{2} \sum_{i\in N}(1 + {\rm log}(\sigma_{i}^2)-\mu_{i}^2-\sigma_{i}^2),
\end{equation}
and
\begin{equation}
{\rm {E_{{\rm reconst}}}} = \frac{1}{N}\sum_{i\in N}\left(y_{{\rm {image}},i}- {y^{\rm G}_{{\rm {image}},i}}\right)^2,
\end{equation}
and
\begin{equation}
\begin{split}
{\rm {E_{{\rm predict}}}} &= \frac{1}{N}\sum_{i\in N}((y_{{\rm {pos}},i}- {y^{\rm G}_{{\rm {pos}},i}})^2 +\\
&(y_{{\rm {ori}},i}- {y^{\rm G}_{{\rm {ori}},i}})^2 + (y_{{\rm {shape}},i}- {y^{\rm G}_{{\rm {shape}},i}})^2),
\end{split}
\label{prediction_error}
\end{equation}
where ${\rm {D_{{\rm KL}}}}$ means KL divergence,~$y^{\rm G}$ is the ground truth output, $N$ is the number of snapshot images.
The hyperparameters $\alpha$ and $\beta$ are used to adjust the loss, with values set to 0.25 and 0.5, respectively, to prioritize the prediction error in~\cref{prediction_error}. 
Due to overlaps between the information necessary for image reconstruction and for predicting object characteristics, we adopt this architecture to efficiently represent it in the latent space. 
Combining multiple errors as shown in~\cref{1st_loss} prevents the over-fitting resulting from only using the prediction error, i.e. the reconstruction error serves as a regularization term which improves convergence properties, leading to smaller number of epochs.
We train this module for 5,000 epochs until the training error converges, using the Adam optimizer~\cite{kingma2014} to update the neural network weights. 
The middle layer of the VAE uses a Sigmoid activation function while the remaining layers use a ReLU activation function.

\subsection{2nd phase haptic-audio and motor module}
The objective of the second phase learning is to predict object characteristics using time-series data from haptic, audio, and motor data, useful for when direct observation of objects is unavailable. 
We record sensorimotor data while commanding the robot to random poses within its workspace, resulting in swinging motions of the container.

The structure of this module comprises an encoder part of VAE and a long short-term memory (LSTM) with a fully connected layer, with the number of neurons detailed in \cref{f:model}.
After training the first phase module, we transfer the latent space value, highlighted in red in~\cref{f:model}, to the second module to use it as the initial value. 
The latent space should encapsulate the information necessary to predict object position, orientation, and shape, which is directly learned from vision. 
Our aim is to leverage this representation to facilitate learning with haptic, audio, and motor data. 
Since the second module handles time-sequential data, we employ LSTM, using sigmoid activation functions and with a sequence of~$T$ time steps, where we iteratively input each single time-step data $x(t)$ (where $t$ denotes the current time step). 
LSTM recurrently learns these inputs and encodes the sequential information in its hidden space $h(t)$ as
\begin{equation}
h(t) = {\rm LSTM}(z(t), z(t-1),...,z(t-T+1)),
\end{equation}
with
\begin{equation}
z(t) = {\rm Encode}(x(t)).
\end{equation}
Then, we use a full-connect layer to compute the position, orientation, and shape from the LSTM hidden latent space as
\begin{equation}
y_{{\rm {pos}}}(t), y_{{\rm {ori}}}(t), y_{{\rm {shape}}}(t) = {\rm FullConnect}(h(t))
\label{eq:2nd_pred}
\end{equation}

The loss function is 
\begin{equation}
\begin{split}
{\rm E} &= \frac{1}{N}\sum_{i\in N}((y_{{\rm {pos}},i}(t)- {y^{\rm G}_{{\rm {pos}},i}}(t))^2 +\\
&(y_{{\rm {ori}},i}(t)-{y^{\rm G}_{{\rm {ori}},i}}(t))^2 + (y_{{\rm {shape}},i}(t)- {y^{\rm G}_{{\rm {shape}},i}}(t))^2),
\end{split}
\label{eq:2nd_loss}
\end{equation}
where here $N$ represents the number of sequential datasets.
We train this module for 20,000 epochs, using the Adam optimizer, until the error converges.

\subsection{Evaluation}
We conducted experiments to assess whether the modules can accurately detect position, orientation, and shape using offline untrained sequential data. 
To evaluate the contribution of cross-model transfer learning, we compared our approach with a baseline where learning starts with haptic, audio, and motor data from scratch, without support from the first phase module. 
Furthermore, we demonstrated object recognition while the robot manipulates the container using both trained and ``untrained'' objects.

\section{Experimental Setup}
\subsection{Hardware design and Objects}
\begin{figure}[t]
    \centerline{\includegraphics[width=8cm]{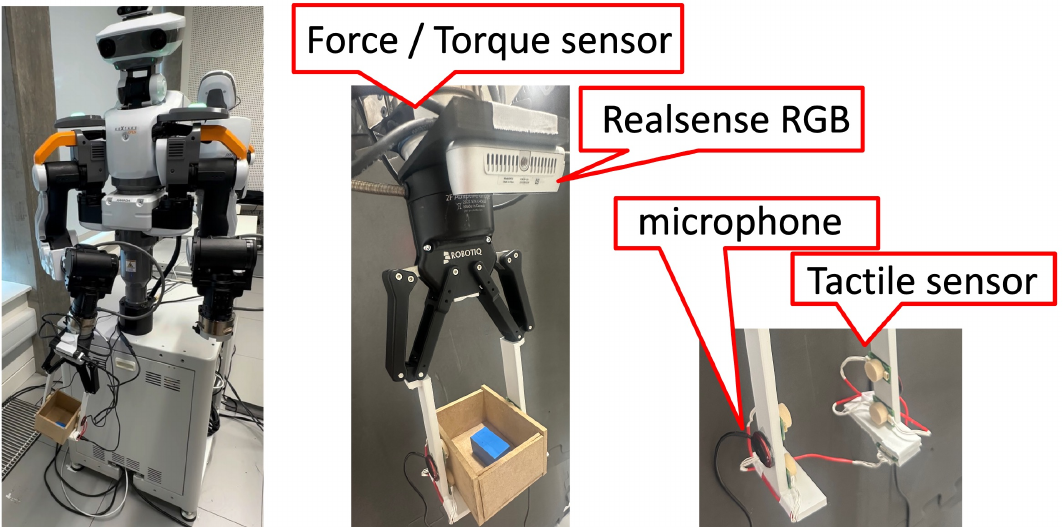}}
    \caption{
       Nextage robot and sensor settings. 
    }
    \label{f:nextage}
\end{figure}

\begin{figure}[t]
    \centerline{\includegraphics[width=8cm]{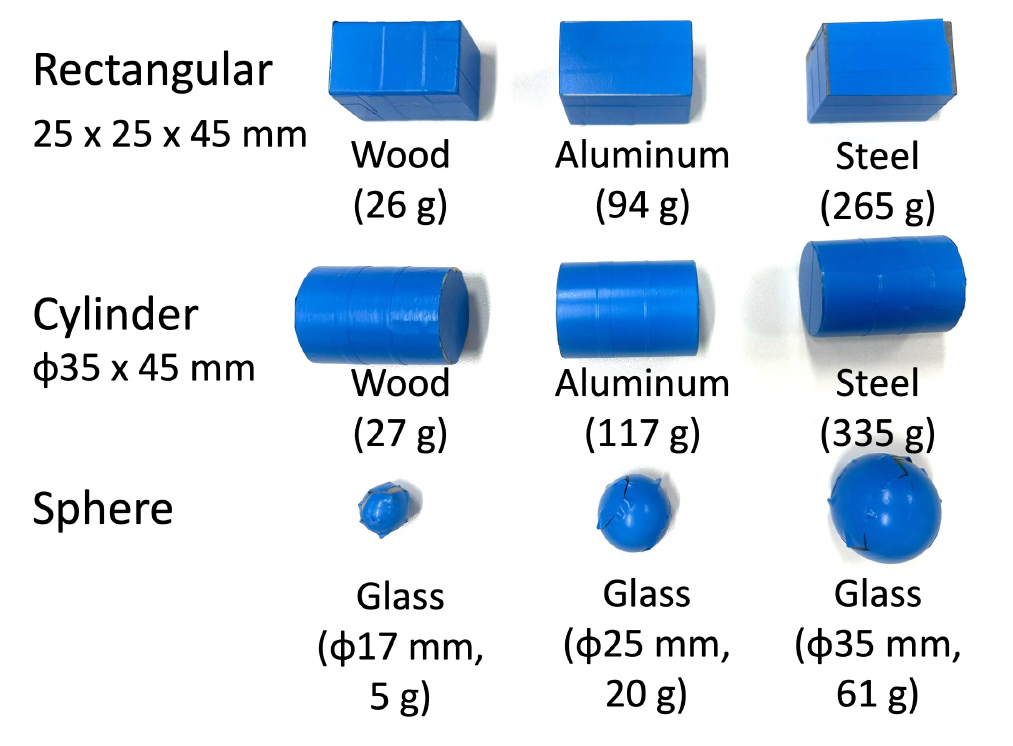}}
    \caption{
       9 different shapes and sizes / weights objects for training.
    }
    \label{f:object}
\end{figure}

\begin{figure*}[t]
    \centerline{\includegraphics[width=16cm]{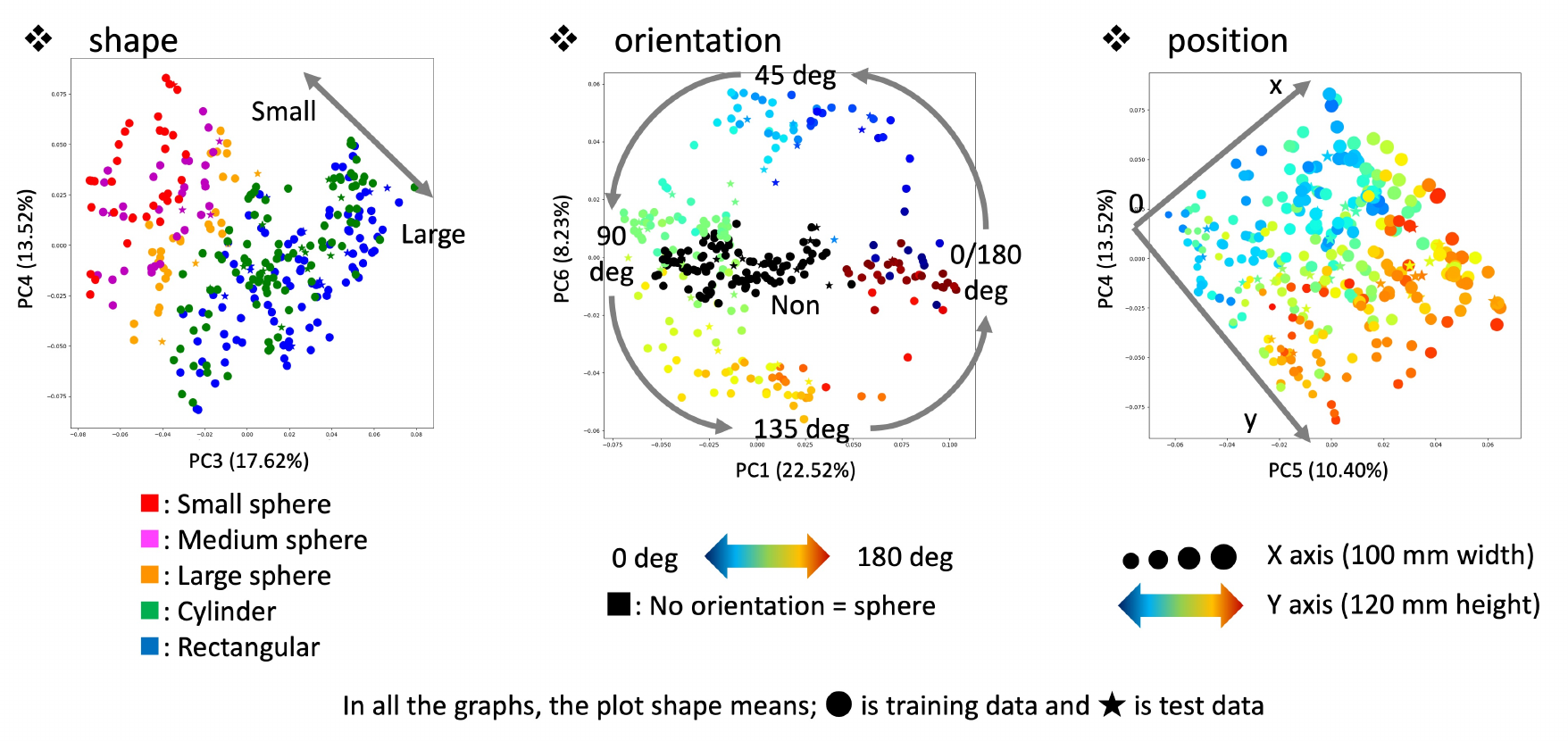}}
    \caption{
       PCA on latent space after training the 1st phase vision module, which is transferred to the 2nd phase haptic-audio module. 
       The latent space represents the object shape, orientation and position.
    }
    \label{f:PCA}
\end{figure*}

\begin{figure}[t]
    \centerline{\includegraphics[width=8cm]{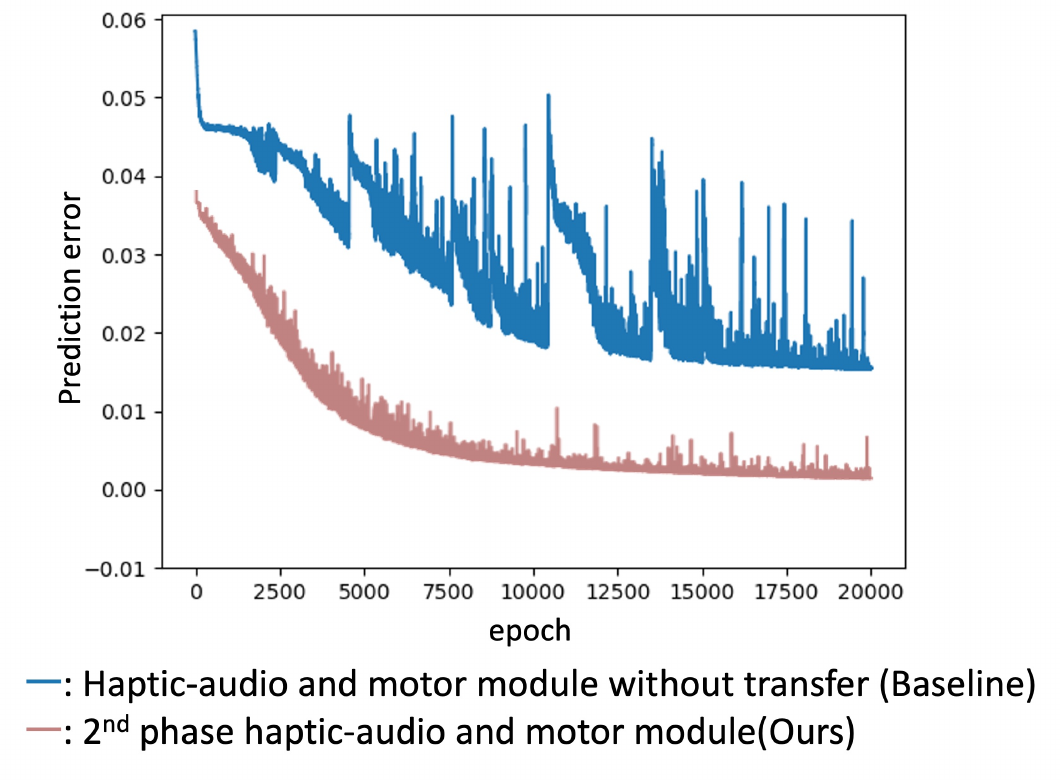}}
    \caption{
       Assessment of prediction error throughout 20,000 epochs of training, contrasting the performance between the proposed method and the baseline approach.
    }
    \label{f:error}
\end{figure}

\begin{figure}[t]
    \centerline{\includegraphics[width=8cm]{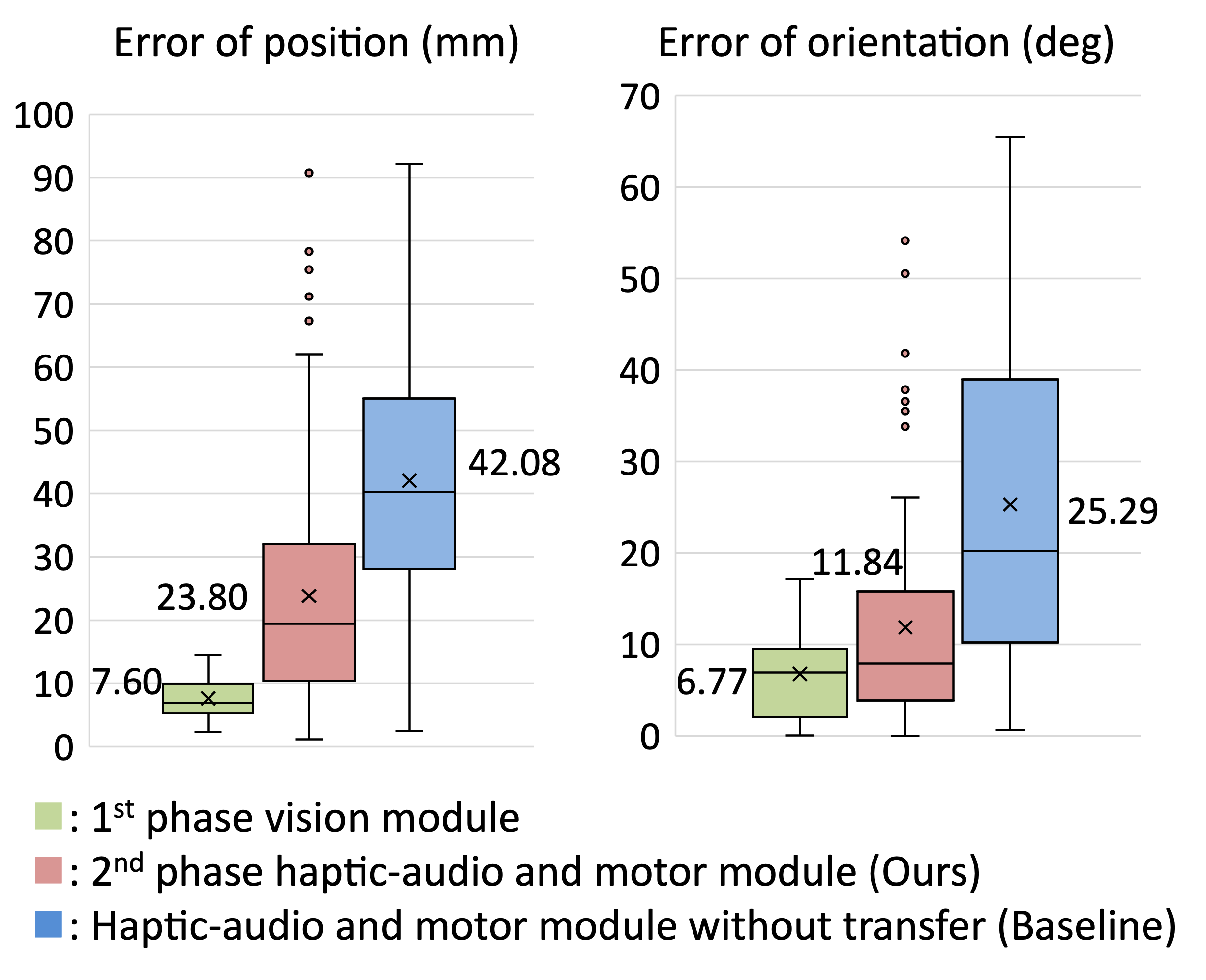}}
    \caption{
       Position and orientation recognition result, showing the prediction error.
       In a box plot, the cross shows the average, the the lower edge of the box represents the first quartile, the line drawn inside the box represents the second quartile, and the upper edge of the box represents the third quartile. 
       The vertical lines represent the maximum and minimum values and the dots are outliers.
    }
    \label{f:box_plot}
\end{figure}

\begin{figure}[t]
    \centerline{\includegraphics[width=9cm]{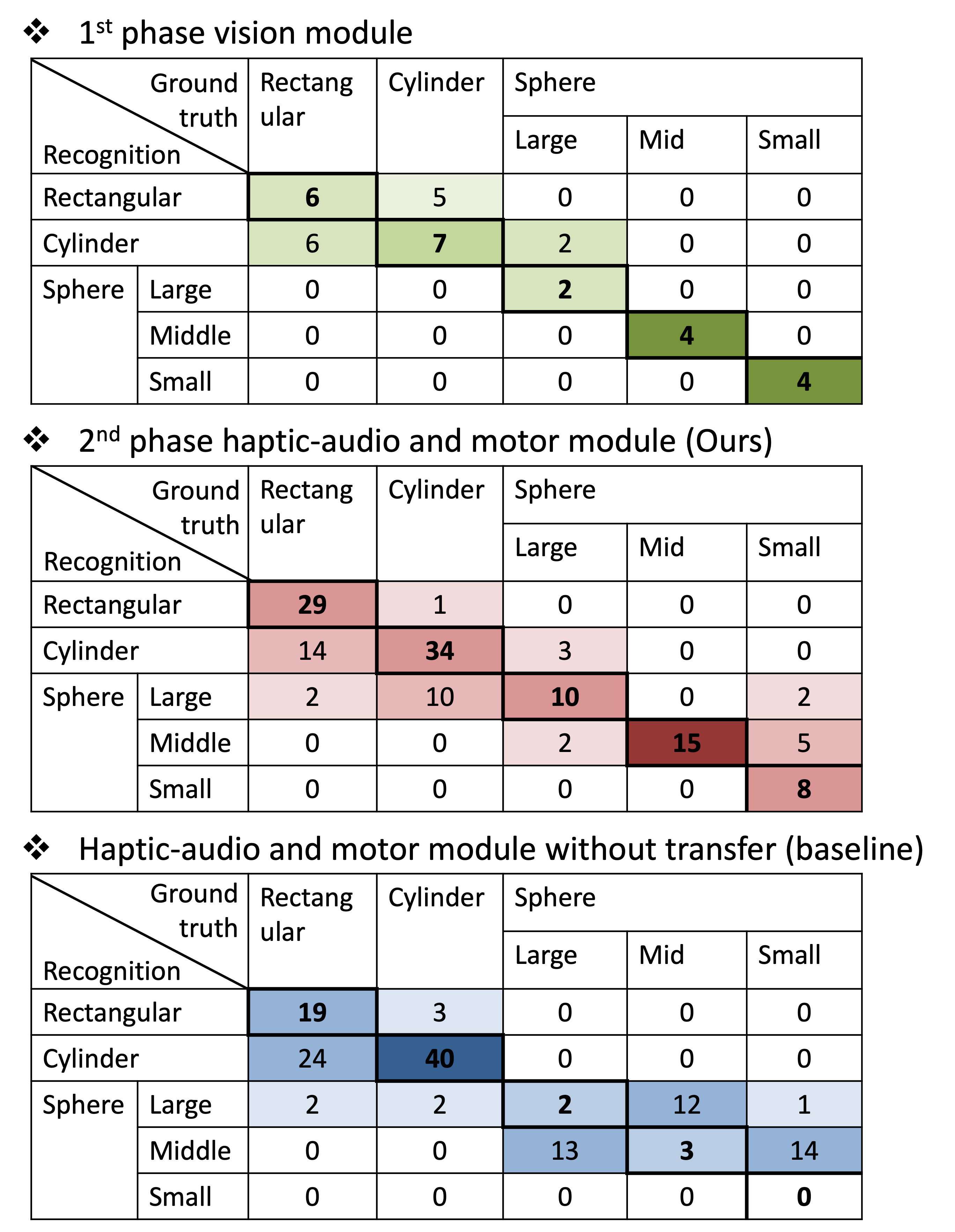}}
    \caption{
       Object shape recognition results which show the number of correct and incorrect predictions.
    }
    \label{f:shape_result}
\end{figure}

\begin{figure}[t]
    \centerline{\includegraphics[width=8.5cm]{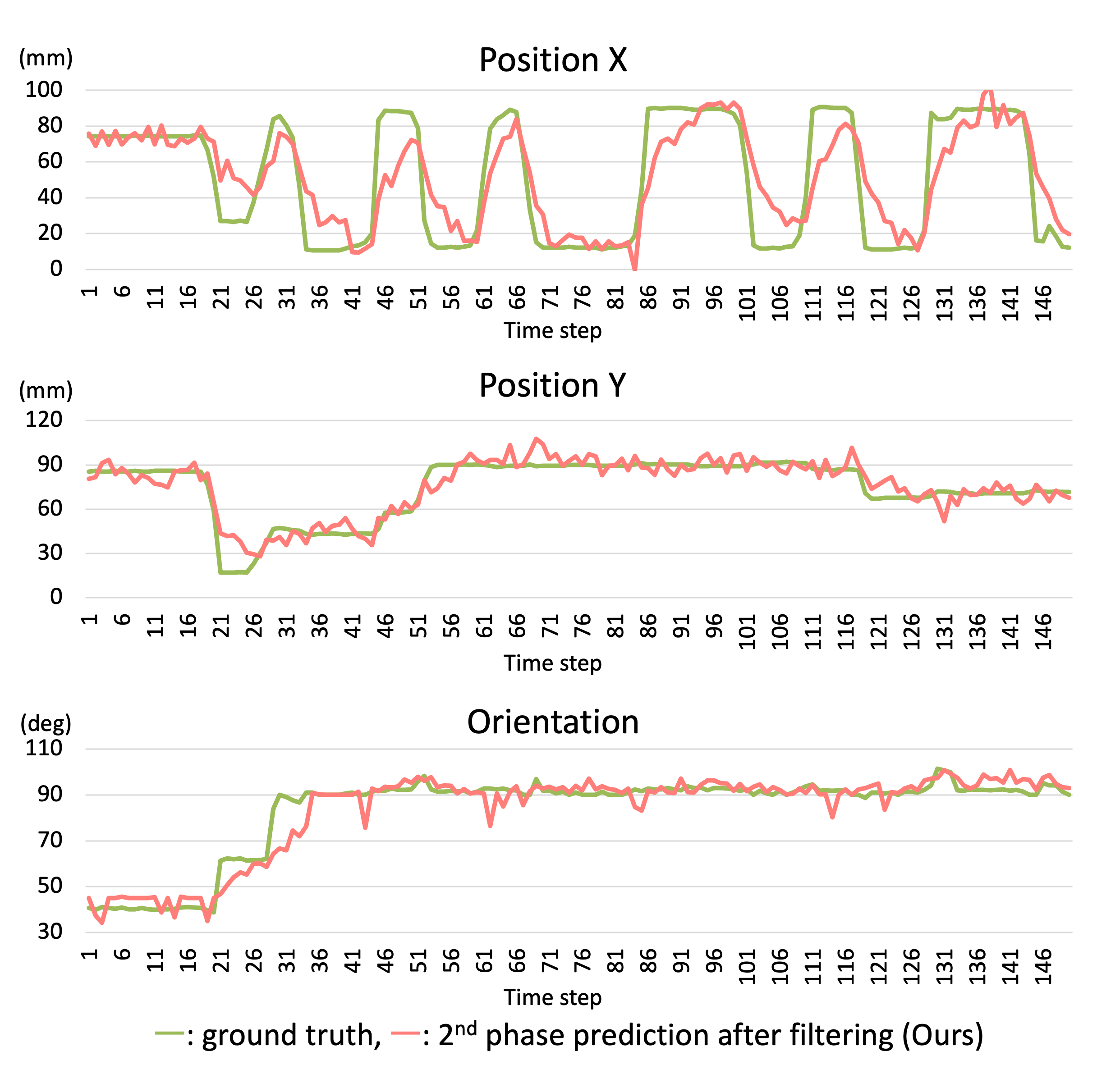}}
    \caption{ 
       Tracking results observed over a 15-second duration (consisting of 150 time steps at a rate of 10 Hz) during the real robot experiment with the wood cylinder and employing the proposed model
    }
    \label{f:smoothing}
\end{figure}

\begin{figure}[t]
    \centerline{\includegraphics[width=8cm]{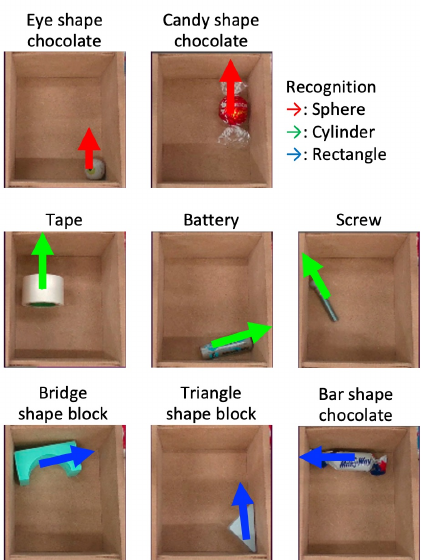}}
    \caption{
       Snapshots of prediction results in the real robot experiment using untrained objects with the proposed model. 
       The arrows are showing the predicted shape with their colour, orientation with their direction, and position with the position they start. 
    }
    \label{f:real_exp}
\end{figure}

\paraDraft{Hardware setting}
In the robot experiment, we utilise the right arm of Nextage Open, which has 6 degrees of freedom (DOF) as depicted in \cref{f:nextage}.
We equip the arm with a Realsense camera D435i, an ATI gamma force and torque sensor, and a Robotiq 2f-140 gripper. 
To enhance gripping stability, we 3D printed gripper fingers to replace the original Robotiq gripper's fingers, are shown in~\cref{f:nextage}. 
Additionally, we integrate four Shokkaku Pot tactile sensors, developed by Touchence Inc., within the gripper fingers and install two piezo contact microphones on the outer sides.

\paraDraft{Object}
We used a box measuring 100 $\times$ 120 $\times$ 70 mm (width $\times$ length $\times$ height), built with 6.3 mm thick wooden boards. 
In training the modules, we utilised 9 different objects of various shapes, sizes, and weights, as summarized in \cref{f:object}. 
To remove any colour-based distinctions in the first vision module, we covered all the objects with the same blue colour.
During the evaluation, we tested 8 untrained objects with characteristics distinct from those of the trained objects.

\subsection{Dataset}
\paraDraft{Data sampling}
We recorded the following data during the experiment:
\begin{itemize}
    \item XYZ position and yaw-pitch-roll of the right arm end-effector (6 dimensions).
    \item RGB images (96 pixels width $\times$ 96 pixels height $\times$ 3 channels).
    \item Force and torque data (6 dimensions).
    \item Tactile sensor data (4 points $\times$ 4 units).
    \item Piezo microphone data (2 units).
\end{itemize}
Prior to inputting the data into the model, we normalize the values of image data to the range [0, 255] and other sensorimotor data to the range [0.2, 0.8]. 

\paraDraft{Data set for 1st}
For the first vision module, we gathered snapshot images with one of the training objects randomly placed inside the box. 
Specifically, we collected 270 images for training the first module, with 30 images captured for each of the 9 training objects. 
Additionally, we acquired 36 images for testing, with 4 images taken for each of the 9 objects.

\paraDraft{Data set for 2nd}
For the second haptic-audio and motor module, we commanded the right arm of the robot to random locations and orientations within its workspace. 
As the robot swung the box randomly, we recorded sequential data from tactile sensors, force-torque sensors, microphones, and end-effector configurations at a frequency of 50 Hz. 
We collected 72 sequence datasets, each comprising 1000 time steps, with 8 sequences captured for each of the 9 objects.
During the training of the second module, we configured the time window to 250 steps, equivalent to 5 seconds, and slid it with a stride of 50 time steps. 
In other words, we trained the model with a total of 1,152 sequential datasets (16 windows $\times$ 72 sequences).
For testing the second module, we utilised 135 different sequential datasets, with 15 datasets recorded for each of the 9 objects.

\subsection{Specification of Ground Truth for Recognition}
\paraDraft{shape}
For object shape, we assigned labels to each dataset using a single dimension as 0.2 for the rectangular prism, 0.35 for the cylinder, 0.5 for the large sphere, 0.65 for the medium sphere, and 0.8 for the small sphere.
We trained the modules to predict these labelled values. 
For classifying the object shapes, we set the classification threshold at the equidistant value between the numerical labels.
Then, we evaluated the classification success based on the number of matches between the predictions and the ground truth shapes.

\paraDraft{Position and orientation}
To obtain the ground truth object positions and orientations, we utilised OpenCV, an open-source computer vision library, to detect blue square or circular shapes in camera images. 
The position contains 2 dimensions: the x and y coordinates of the object's centre.
For setting the ground truth orientation, we initially detected the degree of inclination ($\theta$) of the long edge in the images. 
However, expressing the orientation with the original degree data posed a problem: the actual object orientation is similar between 0 and 179 degrees, yet the values are the most distant. 
To address this issue, we defined the ground truth orientation with 2 dimensions: $(\sin(2\theta), \cos(2\theta))$. This representation enables us to express how close the orientation is to vertical, horizontal, and 45-degree directions clockwise and counterclockwise.
For spheres, which have no inherent orientation, we set both orientation dimensions to 0.

\section{Results and Discussion}
\subsection{Latent space representation} \label{sec:latent_representation}
To analyse the latent space of the VAE, transferred from the first module to the second module, and assess its representation of object characteristics, we applied Principal Component Analysis (PCA) to its latent space. 
\cref{f:PCA} shows a clear distribution of the objects characteristics along the principal components.
Observing the PC3 and PC4 space, we observe that object shapes are ordered according to the size of the objects. 
Additionally, on PC1 and PC6, the representation of orientation demonstrates degrees in a circular manner, with spheres centred at (0, 0) value. 
Furthermore, PC1 primarily indicates whether the orientation is vertical or horizontal, while PC6 expresses clockwise or counterclockwise inclination.
Finally, PC4 and PC5 express the x and y positions in the box.
In conclusion, the latent space effectively distributes and represents geometric and spatial information of the object, which can potentially provide valuable support for the second phase learning.

\subsection{Prediction error convergence during training}
To evaluate whether our proposed transfer method improves the second phase learning, we compared the prediction error calculated with \cref{eq:2nd_loss} with a baseline approach that uses the same haptic-audio and motor learning, without the warm start from the vision module. 
\cref{f:error} shows the convergence of the error during training over 20,000 epochs, where it is evident that the proposed method starts with a lower error than the baseline from the beginning. 
Furthermore, while the prediction error of the baseline exhibits significant fluctuations, the proposed method converges smoothly to a lower level of error. 
Therefore, we conclude that the transfer approach can effectively support training by significantly decreasing the prediction error of object characteristics.

\subsection{Prediction result with test data}
\paraDraft{position and orientation}
\cref{f:box_plot} shows the prediction error of position and orientation, comparing the first phase vision module, our proposal, and the baseline, using the offline dataset.
With our proposal, the average error was 23.80 mm for position and 11.84 degrees for orientation. 
Although these errors are significantly larger when compared with using vision for prediction, that is a much simpler problem setting because the position and orientation are directly observable in that case.
In the baseline, without using vision, the average error was 42.08 mm for position and 25.29 degrees for orientation, with much larger variance. 
This indicates that our proposal contributes to the accuracy of predicting position and orientation, when only using indirect observations.

\paraDraft{Shape}
The tables in \cref{f:shape_result} display the results of shape recognition, where the rows represent the actual shape and the columns represent predictions.
For the first phase vision module, shape recognition is based on direct observation with camera images. 
The success rate of recognition was 63.9\% (23/36 samples), which is relatively low due to the similarity in size between rectangular prisms and cylinders, making them difficult to distinguish from image snapshots.
Conversely, both our proposed model and the baseline use sequential sensorimotor data to predict shape, without relying on vision.
The success rate of our proposed model was 71.1\% (96/135 samples), while the baseline achieved a success rate of 47.4\% (64/135 samples).
As a result, our proposal exhibited the highest accuracy in shape recognition. 
This suggests that the second module could leverage the first phase and further improve shape recognition.

\subsection{Robot experiment with trained and untrained objects}
\label{real_robot}
\paraDraft{Training objects}
We conducted experiments by commanding the robot to swing the box and recognise the object characteristics online, outputting data at a rate of 10 Hz.
With our proposal, predictions ($y_{{\rm {pos}}}(t)$, $y_{{\rm {ori}}}(t)$, $y_{{\rm {shape}}}(t)$) calculated using~\cref{eq:2nd_pred} are independent of previous predictions ($y_{{\rm {pos}}}(t-n)$, $y_{{\rm {ori}}}(t-n)$, $y_{{\rm {shape}}}(t-n)$). 
Consequently, the prediction can be noisy including outliers and exhibit abrupt changes, making it unsuitable for tracking.
To address this issue, we applied a low-pass filter to the output of the model as:
\begin{equation}
    \tilde{y}(t) = 0.4\cdot y(t) + 0.3\cdot\tilde{y}(t-1) + 0.2 \cdot\tilde{y}(t-2) + 0.1 \cdot\tilde{y}(t-3),
\end{equation}
where $\tilde{y}(t)$ is the filtered output.

\paraDraft{Training objects}
\cref{f:smoothing} illustrates the tracking results for 150 time steps, which is a 15-second period, with predictions after passing through the smoothing filter.
Additionally, we provide the attached video demonstrating the tracking results after applying the smoothing filter to the model's predictions.

\paraDraft{Untrained objects}
We further evaluated our model using eight untrained objects, as shown in \cref{f:real_exp}. 
Despite their characteristics differing from the trained objects, the proposed model demonstrated the ability to recognise their shape, position, and orientation. 
The model exhibited generalization capabilities, indicating its potential applicability to a broader range of objects.
The attached video showcases the tracking results after applying the smoothing filter to the predictions for these untrained objects, further demonstrating the effectiveness and generalization of our proposed model.

\section{Conclusion}
\paraDraft{Summary}
In conclusion, this study presents a novel approach to hidden object characteristics recognition in robotic manipulation tasks. 
By employing a two-phase cross-modal transfer learning method, we demonstrate the effectiveness of leveraging visual information from the first phase to improve recognition accuracy with haptic-audio and motor data in the second phase. 
Our experiments show that the proposed method outperforms the baseline approach, achieving better accuracy in predicting object position, orientation, and shape. 
Furthermore, our model exhibits generalization capabilities, successfully recognising untrained objects with characteristics similar to the trained categories. 
These results highlight the potential of our approach in enhancing robotic perception and manipulation in real-world scenarios.

Recognizing characteristics using indirect observations is challenging.
In this work we selected characteristics such as shape, position and orientation because first we could generate a ground truth references for validation and second because the first vision module can directly observe and encode them in its latent space representation.
However, there are other object types of characteristics that also influence the object motion behaviour, such as stiffness, friction, and even deformability, that might have less reliable direct observation from the vision module and make it difficult to generate ground truth references for validation.
Our framework can, in principle, handle these other types of characteristics but that remains untested.

\paraDraft{Next task}
Despite the promising results achieved in this study, there remain areas for improvement. 
As noted in \cref{real_robot}, our current model lacks the incorporation of previous prediction information, which deviates from physical principles. 
Consequently, our model struggles to handle rigid objects remaining motionless and requires five-second time windows and relatively large swinging-box movements to hit the object against a wall at least once for localization.
To address this limitation, future work will focus on implementing a Markov model to assess the reliability of output from the learning model based on previous predictions and output the modified recognition result. 
By incorporating this information, we aim to enhance the precision of predicting the next position and orientation of objects. 

\paraDraft{Future work}
In future work, our focus will shift towards integrating the object characteristics information recognised by our proposed model into robot control systems. 
Specifically, we plan to develop a control system capable of adjusting speed, acceleration, and trajectory based on object characteristics from our recognition model.
By doing so, we aim to enable robots to perform stable and efficient transportation tasks, ensuring that they can interact with objects in a quick and reliable manner in the context of various applications, including logistics, manufacturing, and the service industries.


\printbibliography

\end{document}